\documentclass[letterpaper]{article} % DO NOT CHANGE THIS
\usepackage{aaai24}  % DO NOT CHANGE THIS
\usepackage{times}  % DO NOT CHANGE THIS
\usepackage{helvet}  % DO NOT CHANGE THIS
\usepackage{courier}  % DO NOT CHANGE THIS
\usepackage[hyphens]{url}  % DO NOT CHANGE THIS
\usepackage{graphicx} % DO NOT CHANGE THIS
\usepackage{natbib}  % DO NOT CHANGE THIS AND DO NOT ADD ANY OPTIONS TO IT
\usepackage{caption} % DO NOT CHANGE THIS AND DO NOT ADD ANY OPTIONS TO IT
\usepackage{algorithm}
\usepackage{algorithmic}

\usepackage{newfloat}
\usepackage{listings}
\DeclareCaptionStyle{ruled}{labelfont=normalfont,labelsep=colon,strut=off} % DO NOT CHANGE THIS
\floatstyle{ruled}
\newfloat{listing}{tb}{lst}{}
\floatname{listing}{Listing}
\nocopyright
\usepackage{soul}
\usepackage{xcolor}
\usepackage{multirow}
\usepackage{booktabs}
\usepackage{amsmath}
\usepackage{tikz}
\usetikzlibrary{shapes,arrows.meta,fit}

\title{Faster Inference of Integer SWIN Transformer by Removing the GELU Activation}
\author{
    Mohammadreza Tayaranian\footnote{Correspondence to\\ mohammadreza.tayaranian@mail.mcgill.ca}, Seyyed Hasan Mozafari,\\James J. Clark, Brett Meyer, Warren Gross\\
}
\affiliations{
    Department of Electrical and Computer Engineering\\
    McGill University \\
    Montreal, Canada \\
}

\begin{document}

\maketitle
\begin{abstract}
SWIN transformer is a prominent vision transformer model that has state-of-the-art accuracy in image classification tasks.
Despite this success, its unique architecture causes slower inference compared with similar deep neural networks.
Integer quantization of the model is one of the methods used to improve its inference latency.
However, state-of-the-art has not been able to fully quantize the model. 
In this work, we improve upon the inference latency of the state-of-the-art methods by removing the floating-point operations, which are associated with the GELU activation in Swin Transformer.
While previous work proposed to replace the non-integer operations with linear approximation functions, we propose to replace GELU with ReLU activation.
The advantage of ReLU over previous methods is its low memory and computation complexity.
We use iterative knowledge distillation to compensate for the lost accuracy due to replacing GELU with ReLU.
We quantize our GELU-less SWIN transformer and show that on an RTX 4090 NVIDIA GPU we can improve the inference latency of the quantized SWIN transformer by at least $11\%$ while maintaining an accuracy drop of under $0.5\%$ on the ImageNet evaluation dataset.
\end{abstract}

\section{Introduction}

The attention mechanism has gained popularity in recent years after its successful debut in transformer architecture \cite{vaswani2017attention}.
While the transformer architecture has been initially used for natural language processing (NLP) tasks, it was also brought to the computer vision domain with the introduction of vision transformer models \cite{dosovitskiyimage}.
SWIN transformer \cite{swin} is a well-known vision transformer which improves on the original design by using shifted windows in the input.
It shows state-of-the-art performance in a variety of computer vision tasks.
However, SWIN transformer's inference latency is negatively affected due to its use of windowed attention.
The windowed attention relies on shifting of the input activations, and the shift operation is highly memory intensive, thus having a high impact on the inference latency.
For instance, running inference on an NVIDIA V100 GPU, SWIN\textsubscript{\textrm{SMALL}} is shown to be $55\%$ slower compared to ViT\textsubscript{\textrm{SMALL}} \cite{liu2022convnet}.
For mobile devices, \citet{wang2022towards} demonstrated a more pronounced gap in the inference latency where SWIN\textsubscript{\textrm{SMALL}} is $2.2$ times slower than ViT\textsubscript{\textrm{SMALL}}.

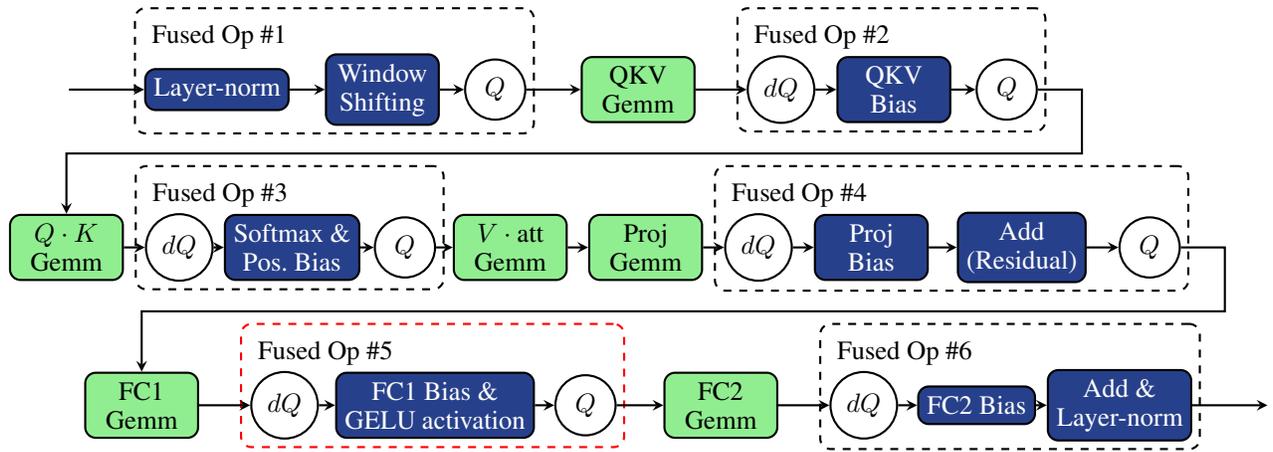
\begin{figure*}[t]
    \centering
    \begin{tikzpicture}[>=stealth, node distance=1.5cm]

\definecolor{darkblue}{RGB}{39,64,139}
\definecolor{lightgreen}{RGB}{144,238,144}

% Define styles for rectangles and circles
\tikzstyle{rect} = [rectangle, draw=black, thick, rounded corners, minimum width=1.5cm, minimum height=0.5cm, text centered, fill=darkblue, text=white, align=center]
\tikzstyle{rect2} = [rectangle, draw=black, thick, rounded corners, minimum width=1.5cm, minimum height=0.5cm, text centered, fill=lightgreen, text=black, align=center]
\tikzstyle{rectd} = [rectangle, draw=black, dashed, thick, rounded corners, minimum width=1.5cm, minimum height=0.5cm, text centered, inner sep=0.1cm]
\tikzstyle{circ} = [circle, draw=black, thick, minimum size=0.8cm, text centered]

% Draw shapes
\node (a1) [rect] {Layer-norm};
\node (a2) [rect, right of=a1, xshift=0.7cm] {Window\\Shifting};
\node (a3) [circ, right of=a2] {$Q$};
\node (ida) [at=(a1.west), yshift=0.7cm, xshift=1.0cm, align=right] {Fused Op \#1};
\node (rect1) [rectd, fit=(a1)(a2)(a3)(ida)] {};

\node (b) [rect2, right of=a3, xshift=0.4cm] {QKV\\Gemm};

\node (c1) [circ, right of = b, xshift=0.4cm] {$dQ$};
\node (c2) [rect, right of=c1] {QKV\\Bias};
\node (c3) [circ, right of=c2] {$Q$};
\node (idc) [at=(c1.west), yshift=0.7cm, xshift=1.0cm, align=left] {Fused Op \#2};
\node (rect2) [rectd, fit=(c1)(c2)(c3)(idc)] {};

\node (d1) [rect2, below of=a1, xshift=-2cm, yshift=-0.6cm] {$Q \cdot K$\\Gemm};

\node (d2) [circ, right of=d1] {$dQ$};
\node (d3) [rect, right of=d2] {Softmax \&\\Pos. Bias};
\node (d4) [circ, right of=d3] {$Q$};
\node (idd) [at=(d2.west), yshift=0.7cm, xshift=1.0cm, align=left] {Fused Op \#3};
\node (rect3) [rectd, fit=(d2)(d3)(d4)(idd)] {};

\node (d5) [rect2, right of=d4, xshift = -0.1cm] {$V \cdot \textrm{att}$\\Gemm};

\node (e1) [rect2, right of=d5, xshift=0.3cm] {Proj\\Gemm};

\node (f1) [circ, right of=e1] {$dQ$};
\node (f2) [rect, right of=f1] {Proj\\Bias};
\node (f3) [rect, right of=f2, xshift=0.5cm] {Add\\(Residual)};
\node (f4) [circ, right of=f3, xshift=0.2cm] {$Q$};
\node (idf) [at=(f1.west), yshift=0.7cm, xshift=1.0cm, align=left] {Fused Op \#4};
\node (rect4) [rectd, fit=(f1)(f2)(f3)(f4)(idf)] {};

\node (f) [rect2, below of=d1, xshift=1cm, yshift=-0.6cm] {FC1\\Gemm};

\node (g1) [circ, right of = f, xshift=0.4cm] {$dQ$};
\node (g2) [rect, right of=g1, xshift=0.5cm] {FC1 Bias \& \\GELU activation};
\node (g3) [circ, right of=g2, xshift=0.5cm] {$Q$};
\node (idg) [at=(g1.west), yshift=0.7cm, xshift=1.0cm, align=left] {Fused Op \#5};
\node (rect5) [rectd, draw=red, fit=(g1)(g2)(g3)(idg)] {};

\node (h) [rect2, right of=g3, xshift=0.3cm] {FC2\\Gemm};

\node (i1) [circ, right of=h, xshift=0.4cm] {$dQ$};
\node (i2) [rect, right of=i1, xshift=0cm] {FC2 Bias};
\node (i3) [rect, right of=i2, xshift=0.4cm] {Add \&\\Layer-norm};
\node (idi) [at=(i1.west), yshift=0.7cm, xshift=1.0cm, align=left] {Fused Op \#6};;
\node (rect6) [rectd, fit=(i1)(i2)(i3)(idi)] {};

% Straight arrows
\draw[->, thick] (a1) -- (a2);
\draw[->, thick] (a2) -- (a3);
\draw[->, thick] (a3) -- (b);
\draw[->, thick] (b) -- (c1);
\draw[->, thick] (c1) -- (c2);
\draw[->, thick] (c2) -- (c3);

\draw[->, thick] (d1) -- (d2);
\draw[->, thick] (d2) -- (d3);
\draw[->, thick] (d3) -- (d4);
\draw[->, thick] (d4) -- (d5);
\draw[->, thick] (d5) -- (e1);
\draw[->, thick] (e1) -- (f1);
\draw[->, thick] (f1) -- (f2);
\draw[->, thick] (f2) -- (f3);
\draw[->, thick] (f3) -- (f4);

\draw[->, thick] (f) -- (g1);
\draw[->, thick] (g1) -- (g2);
\draw[->, thick] (g2) -- (g3);
\draw[->, thick] (g3) -- (h);
\draw[->, thick] (h) -- (i1);
\draw[->, thick] (i1) -- (i2);
\draw[->, thick] (i2) -- (i3);

% bended lines
\draw[thick,->] (c3) -- ++(1,0) -- ++(0,-0.85) -- ++(-13.5,0) -- (d1.north);
\draw[thick,->] (f4) -- ++(1,0) -- ++(0,-0.85) -- ++(-14.4,0) -- (f.north);

% start and finish line

\draw[->, thick] ([xshift=-1cm]a1.west) -- ++(1,0);
\draw[->, thick] (i3.east) -- ++(1,0);

\end{tikzpicture}
    \caption{
        High-level depiction of the components of a transformer block in the quantized SWIN-transformer.
        $Q$ and $dQ$ denote the quantization and de-quantization operations, respectively.
    }
    \label{fig:overall}
\end{figure*}

Quantization is one of the techniques used for the improvement of the inference latency of deep neural networks.
It involves representing the values of the neural network using data types with lower bit-widths.

Despite the theoretical possibility of using arbitrary data types and bit-widths for quantization, the achieved inference speedup depends on hardware on which the model is being deployed \cite{vaqf}.
For instance, consider a quantization method which uses the 4-bit integer data type to represent weights and activations of a deep neural network.
Running this method on hardware that doesn't support 4-bit arithmetic operations results in a speedup lower than the expected theoretical speedup.
Besides, the overhead of converting the quantized values to a data type which is supported by the GPU will further reduce the speedup of the method.

Another important factor in the speedup of partially quantization models is their non-integer components~\cite{ivit}.
Such component is the non-linear function, e.g. Softmax, that due to its non-linearity is not easily quantizable.
Given a non-linear function $f$, its quantized input $\hat{x}$, and the quantization scale $s$, we have $f(s \hat{x}) \neq s f(\hat{x})$.
As a result, some integer implementations opt to use the floating-point data type for these components.
This enforces the inclusion of memory intensive quantization and de-quantization functions in the inference pipeline which results in notable overhead.

With the goal of avoiding non-integer components, previous work focused on quantizing the non-linear operations of transformer-based models \cite{ivit, ibert, lin2021fq}.
The main theme of these works is to substitute the non-linear components with a linear or piece-wise linear version without losing accuracy.
In this work, we propose to replace the GELU activation with the piece-wise linear ReLU function \cite{fukushima1975cognitron}.
Compared with which needs to compute the maximum of the input tensor to approximate GELU \cite{ivit}, ReLU can be simply applied with the help of a comparator.
The advantage of ReLU is its low complexity and simple logic whereas previous work's shift-based GELU required  \cite{ivit}.
We apply these changes to the SWIN transformer model in a layer-by-layer fashion and use knowledge distillation in the process to maintain the model's accuracy.
The weights and input activations of the resulting model, which we call GELU-less SWIN, are then quantized using post-training quantization.

The results of this comparison show that our model has a maximum accuracy drop of $0.5\%$ while achieving more than  $11\%$ inference latency reduction compared to the FasterTransformer framework.

\section{Previous Work}

\subsection{Quantization of Linear Components}

The majority of previous work focuses on the quantization of linear components of vision transformers, i.e. fully connected and convolutional layers.
In the quantized version of these components, either the weight, the input activation, or both matrices are quantized.
The quantization scale is obtained by either quantization aware training or via a calibration phase in a post-training quantization fashion.

\citet{vit-ptq} propose a mixed-precision quantization in which each linear layer in different transformer blocks has a different bit-width.
\citet{ptq4vit} uses a Hessian-guided similarity measure for finding quantization scales.
To keep the precision in the layers that are more sensitive to the quantization noise, they use two scale factors for each fully connected layer with each of the scales responsible for only a part of the tensor.
\citet{qvit1} quantizes the linear layers of vision transformers down to 2 bits.
They do so by adding trainable parameters that help the quantized weights follow the distribution of floating-point weights.
\citet{qvit2} uses different bit-widths and scales for each attention head.
\citet{qvit1} and \citet{qvit2} both show promising results in terms of accuracy when using 4-bit and 3-bit weights.
Despite being able to maintain the model's test accuracy, all of these works lack studies of hardware performance metrics, e.g. latency, of their quantized model and only discuss the model size, which is not a reliable proxy for latency. 
In the present work, in addition to accuracy, we measure the inference latency of our quantization method and its speedup compared to the baseline.

\subsection{Quantization of Non-Linear Components}

Another line of work tries to quantize the non-linear components of the model.
Softmax, LayerNorm, and GELU activation are the three main non-linear components of vision transformers that are not straightforward to quantize.

\citet{lin2020towards} and \citet{ibert} use polynomial approximations to provide quantizable versions of the non-linear components.
Although their methods were proven successful for transformer-based language models, \citet{ivit} have shown that these methods cannot be used to vision transformers.
\citet{lin2021fq} propose a $\log_2$ quantization method which adapts the methodology of \cite{ibert} for fully integer vision transformers.
Although the authors show their method's ability to maintain accuracy, the performance of their method in terms of hardware metrics like latency is not discussed.

The closest work to our work is I-ViT, which provides shift-based replacements for the non-linear components \cite{ivit}.
Their integer-friendly replacement functions use the power of two approximations of the $e^x$ function \cite{stevens2021softermax}.
They also improve on the integer LayerNorm proposed by \citet{lin2020towards} by using a shift-based iterative function to compute the square root of the variance.
Their experimental results show that their quantized model has a $5.8\%$ shorter inference latency compared with implementing quantized SWIN transformer based on NVIDIA's FasterTransformer framework.
Despite its high accuracy, their proposed shift-based GELU approximation has a high memory and computation complexity as it needs to compute the maximum value of the input tensor to approximate GELU.
In comparison, our proposed method of replacing the GELU activation with ReLU has the advantage of lower memory and computation complexity given the ReLU's simple logic.
\section{Background and Motivation}

SWIN transformer addresses an architectural problem with the original vision transformer model.
It uses a windowed attention mechanism to avoid global attention and its considerable computation.
The window attention divides each input activation into smaller windows and computes the attention over the image patches in each window.
This enables the use of SWIN transformer as a backbone model for tasks such as semantic segmentation that have larger input images \cite{swin}.

Despite its state-of-the-art accuracy, SWIN transformer's use of window shifting operations has negative effects on its hardware performance.
This negative effect is revealed when comparing the inference latency of SWIN transformer with the original vision transformer.
For instance, SWIN\textsubscript{\textrm{SMALL}} is $55\%$ slower in terms of inference latency when compared to DeiT\textsubscript{\textrm{SMALL}} which has the same architecture as the original vision transformer \cite{liu2022convnet}.
In the case of mobile GPU, \citet{mehta2021mobilevit} demonstrated that the window shifting operations are not supported by iPhone GPUs, making it impossible to implement SWIN on this hardware.
Nevertheless, on mobile GPUs where SWIN can be implemented, SWIN\textsubscript{\textrm{SMALL}} is $2.2$ times slower than DeiT\textsubscript{\textrm{SMALL}} \cite{wang2022towards}.

\begin{table}[t]
    \centering
    \begin{tabular}{c|cccccc}
     & \multicolumn{6}{|c}{Fused Op \#} \\
     Model & 1 & 2 & 3 & 4 & 5 & 6 \\
    \cmidrule[1.2pt](l{\abovetopsep}r{\belowbottomsep}){1-7}
    SWIN\textsubscript{\textrm{TINY}} & $0.32$ & $1.51$ & $2.55$ & $0.88$ & $2.03$ & $1.16$ \\
    SWIN\textsubscript{\textrm{SMALL}} & $0.3$ & $2.01$ & $4.1$ & $1.17$ & $2.48$ & $1.36$ \\
    SWIN\textsubscript{\textrm{BASE}} & $0.57$ & $2.53$ & $5.44$ & $1.57$ & $3.89$ & $2.69$ \\
    SWIN\textsubscript{\textrm{LARGE}} & $1.22$ & $3.93$ & $8.18$ & $2.92$ & $8.13$ & $3.66$ \\
    \midrule
    Average & $0.6$ & $2.5$ & $5.07$ & $1.64$ & $4.13$ & $2.22$ \\        
\end{tabular}
    \caption{
    Latency (ms) of the fused operations of the quantized SWIN transformer, depicted in Figure \ref{fig:overall}.
    The latency value are measured on an NVIDIA RTX 4090 and are averaged over 1000 inference runs.
    }
    \label{tab:layer_by_layer}
\end{table}

Inspired by the gap in the inference latency, we use integer quantization to speed up the inference of the SWIN transformer model.
Since our target hardware is the NVIDIA GPU, we start with the FasterTransformer framework's proposed quantized SWIN.
Figure \ref{fig:overall} depicts a high-level overview of the components of this quantized SWIN transformer.
$Q$ and $dQ$ are the quantization and de-quantization functions that are used to convert between the integer and floating-point data types.
The quantized SWIN uses 8-bit integer for the weights and input activations of the linear layers.
It also uses GPU's integer tensor cores to accelerate the integer matrix multiplication operations.

In this quantized SWIN architecture, components like biases, Softmax, layer-norms, and the residual connections use the floating-point data type.
The quantized SWIN uses fused operations, shown in Figure \ref{fig:overall} with dashed rectangles, to spread the overhead of quantization and de-quantization of the integer values over multiple functions.
The functions inside each fused operation use the \textit{shared memory} of the GPU to pass values between each other.
This design minimizes the accesses to the slow \textit{global memory} and thus keeps the latency of the fused operation, and the entire model, at a minimum.

\section{GELU-less SWIN Transformer}

The quantized SWIN transformer is depicted in Figure \ref{fig:overall}.
The components that are using the 8-bit integer GEMM, shown in light green, are already quantized to the minimum bit-width supported by the GPU.
Thus, we turn our attention to the fused operations and measure their latencies.

The latency of each fused operation, calculated as the drop in inference latency resulted from removing it, is provided in Table \ref{tab:layer_by_layer}.
Based on these results, Softmax and GELU activation are the two non-integer components that have the highest latency in the quantized SWIN transformer pipeline.
Furthermore, our experiments reveal that the latency of all the fused operations are dominated by the global memory accesses of the required quantization and de-quantization functions in a not fully quantized implementation.
As a result, instead of modifying only parts of a fused operation, we need to remove it entirely to achieve higher inference speedup.

Considering these observations, we propose to remove the fused operation associated with the GELU activation and substitute it with an integer activation function.
We propose to replace GELU with the piece-wise linear ReLU activation.
While the Shift-based GELU proposed by \citet{ivit} needs to compute the maximum value of the input tensor, the ReLU function is a simple activation function and thus has a lower memory complexity than its Shift-based alternative.
In order to completely remove the fused operation of GELU, we also need to eliminate the \textit{FC1 Bias} as well.
We avoid changing the Softmax fused operation as it also contains the relative position bias which is essential to converting input image to tokens.

\begin{algorithm}[tb]
\caption{Our proposed GELU replacement method with knowledge distillation.}
\label{alg:algorithm}
\textbf{Input}: SWIN, dataset\\
\textbf{Parameter}: $N$: Number of Transformer Blocks\\
\textbf{Output}: GELU-less SWIN
\begin{algorithmic}[1] %[1] enables line numbers
\STATE $\textrm{student} \gets \operatorname{clone}(\textrm{SWIN})$
\FOR{$i \gets 1$ to $N$}
    \STATE $\textrm{block} \gets \textrm{student}.\textrm{blocks}[i]$
    \STATE $\textrm{block}.\textrm{activation} \gets \operatorname{ReLU}$
    \STATE $\textrm{block}.\textrm{bias} \gets 0$
    \STATE $\operatorname{disable\_gradient} (\textrm{block}.\textrm{bias}) $
    \STATE $\operatorname{kd\_epoch} (\textrm{student}, \textrm{SWIN}) $
    
\ENDFOR
\STATE $\textrm{GELU-less SWIN} \gets \textrm{student}$
\STATE \textbf{return} GELU-less SWIN
\end{algorithmic}

\end{algorithm}

\begin{table*}[t]
    \centering
    
    \begin{tabular}{c|r|cccc}
         % \cline{3-7}
         \multirow{2}{*}{Model} & \multirow{2}{*}{Method} & \multirow{2}{*}{Datatype} & Top-1 Acc. & Latency & \multirow{2}{*}{Speedup} \\
         & & & (\%) & (ms)
         \\
         \midrule
         
         \multirow{4}{*}{SWIN\textsubscript{\textrm{TINY}}}
          & Baseline & FP32 & $81.2$ & $60.27$ & $\times 1$ \\
          & Half-precision & FP16 & $81.2$ & $24.96$ & $\times 2.41$ \\
          & FasterTransformer & int8 & $80.1$ & $17.04$ & $\times 3.54$ \\
          & Ours & int8 & $80.0$ & $15.01$ & $\times \mathbf{4.02}$ \\

          \cmidrule[1.2pt](l{\abovetopsep}r{\belowbottomsep}){1-6}

           \multirow{4}{*}{SWIN\textsubscript{\textrm{SMALL}}}
          & Baseline & FP32 & $83.2$ & $103.21$ & $\times 1$ \\
          & Half-precision & FP16 & $83.2$ & $40.26$ & $\times 2.56$ \\
          & FasterTransformer & int8 & $83.0$ & $25.05$ & $\times 4.12$ \\
          & Ours & int8 & $82.5$ & $22.57$ & $\times \mathbf{4.57}$ \\

          \cmidrule[1.2pt](l{\abovetopsep}r{\belowbottomsep}){1-6}

           \multirow{4}{*}{SWIN\textsubscript{\textrm{BASE}}}
          & Baseline & FP32 & $83.4$ & $157.31$ & $\times 1$ \\
          & Half-precision & FP16 & $83.4$ & $58.39$ & $\times 2.69$ \\
          & FasterTransformer & int8 & $83.3$ & $35.55$ & $\times 4.43$ \\
          & Ours & int8 & $84.6$ & $31.66$ & $\times \mathbf{4.97}$ \\

          \cmidrule[1.2pt](l{\abovetopsep}r{\belowbottomsep}){1-6}

           \multirow{4}{*}{SWIN\textsubscript{\textrm{LARGE}}}
          & Baseline & FP32 & $86.2$ & $284.41$ & $\times 1$ \\
          & Half-precision & FP16 & $86.2$ & $104.76$ & $\times 2.71$ \\
          & FasterTransformer & int8 & $85.8$ & $61.35$ & $\times 4.64$ \\
          & Ours & int8 & $85.5$ & $53.22$ & $\times \mathbf{5.34}$ \\

    \end{tabular}

    \caption{
    Comparison of our proposed method with the floating-point baselines and FasterTransformer's quantizated model.
    The accuracy are from evaluating a pre-trained SWIN transformer on the ImageNet evaluation dataset.
    The reported inference latency is for a batch size of 128, averaged over 1000 runs.
    }
    \label{tab:main}
\end{table*}

Algorithm \ref{alg:algorithm} describes our proposed method for replacing the GELU activation with ReLU.
To avoid risking model divergence, we gradually apply our changes to the model and modify the transformer blocks one at a time.
After each block modification, we use knowledge distillation to distill the soft labels of the fully GELU SWIN to our partially GELU and partially ReLU model.
The result of the algorithm, which we call the GELU-less SWIN, does not have a floating-point fused operation for its activation function.
As ReLU is easily quantizable, it can be fused, as an integer operation, to the previous GEMM.
This way, ReLU's latency will be completely masked as it will be directly applied to the output of the GEMM.
Finally, we apply the post-training quantization method of the FasterTransformer framework to this GELU-less SWIN transformer and quantize its weights and input activations.

\section{Experiments}

\subsection{Experimental Setup}

We study the evaluation accuracy and latency of our method by applying it to variuos configurations of the SWIN transformer.
We use a pre-trained SWIN model which is pre-trained on the ImageNet training dataset and evaluate it on the ImageNet evaluation dataset \cite{deng2009imagenet}.
As our methodology mainly targets NVIDIA GPU hardware, we implement our method on top of NVIDIA's FasterTransformer\footnote{https://github.com/NVIDIA/FasterTransformer} framework.
We compare the performance of our quantized SWIN with FasterTransformer's quantized SWIN and also the 32-bit and 16-bit floating-point models.

For knowledge distillation, we use the SGD optimizer with a constant learning rate of $10^{-2}$ and a momentum of $0.9$.
As described in Algorithm \ref{alg:algorithm}, the number of knowledge distillation epochs is the same as the number of transformer blocks of the SWIN model.
So, SWIN\textsubscript{\textrm{TINY}} undergoes 12 epochs of the knowledge distillation while for other SWIN configurations this number is 24 epochs.
We use $10\%$ of the ImageNet training dataset for our knowledge distillation, with a batch size of $32$.
The wall clock time for the knowledge distillation is $74$, $203$, $213$, and $337$ minutes, respectively, for models from SWIN\textsubscript{\textrm{TINY}} to SWIN\textsubscript{\textrm{LARGE}}.

For our proposed method, we apply the same post-training quantization that is used in the FasterTransformer framework on our GELU-less SWIN.
The inference latency is an average of 1000 runs and is measured on a quantized model which had its GELU fused operation removed.
Our batch size for the evaluation experiments is $128$.
All our experiments are performed on an NVIDIA RTX 4090 GPU.

\subsection{Experimental Results}

Table \ref{tab:main} provides the evaluation top-1 accuracy and latency of SWIN transformer configurations using different methods.
As the Table demonstrates, our proposed GELU-less quantized SWIN has the smallest inference latency across the SWIN configurations.
Our method is able to improve the latency of the FasterTransformer by $12\%$, $11\%$, $11\%$, and $13\%$, respectively, for models from SWIN\textsubscript{\textrm{TINY}} to SWIN\textsubscript{\textrm{LARGE}}.

As an ablation study, we also apply our method without the use of knowledge distillation which results in an evaluation accuracy of less than $0.9\%$ which shows the knowledge distillation is essential to the proposed algorithm.

\section{Conclusion}

In this work we proposed a method to reduce the inference latency of int-8 quantized SWIN transformer model.
We analyzed the latency of the operations in the existing int-8 quantized SWIN piepline.
Based on the analysis, we proposed to replace the floating-point GELU activation with the ReLU activation.
ReLU is a piece-wise linear function which is easily quantizable and has a very low complexity.
Our proposed method replaces GELU with ReLU and removes the bias that is fused to it.
We also use knowledge distillation to maintain the accuracy.
Our experiments show that quantizing our proposed GELU-less SWIN results in at least $11\%$ reduction of inference latency compared to the original quantized SWIN transformer model.

\bibliography{aaai24}
\end{document}